\documentclass[runningheads]{llncs}
\usepackage[T1]{fontenc}
\usepackage{graphicx}
\usepackage{amsmath, amssymb}
\usepackage{subcaption}

\begin{document}
\title{Resource-Constrained Affect Modelling via Variance Regularisation Pruning}
%
%
\author{Kosmas Pinitas\inst{1,2,}
\thanks{Kosmas Pinitas is the corresponding author of this paper}
\and
Konstantinos Katsifis\inst{1,2}
}
\authorrunning{Pinitas et al.}
%
\institute{Mediterranean College, Athens, Greece \and
University of Derby, Derby, UK\\
\email{k.pinitas@mc-class.gr}, \email{k.katsifis@mc-class.gr}
}
\maketitle              
\begin{abstract}
Affective computing systems are increasingly embedded in pervasive and interactive environments, such as adaptive games, assistive technologies, and resource-constrained platforms, where computational efficiency must be balanced with reliability across diverse users. Model pruning offers an effective way to reduce computational demands, yet existing approaches typically optimise for sparsity alone, without accounting for how parameter removal impacts robustness across individuals. In this work, we introduce Variance-Regularised Pruning (VR), a pruning framework that explicitly incorporates cross-participant stability into the sparsification process. Rather than relying solely on average prediction error, VR evaluates each connection based on its joint contribution to both prediction accuracy and variability across users, prioritising parameters that remain reliable under distributional differences. We evaluate the proposed approach on the AGAIN dataset, which includes arousal annotations collected across nine affect-eliciting game environments. Experimental results demonstrate that VR maintains competitive Concordance Correlation Coefficient (CCC) performance even at 80\% sparsity without additional fine-tuning, highlighting its suitability for deployment in real-world, resource-limited affect-aware systems. Overall, the proposed framework supports the development of compact, robust affective models that can operate reliably in real-world interactive environments.

\end{abstract}

\section{Introduction}

Affective computing systems are increasingly deployed in pervasive and interactive environments, where models must operate under resource constraints while delivering reliable predictions across diverse users~\cite{picard1999affective,yannakakis2023affective}. Applications such as emotion-aware interfaces and interactive media rely on consistent affect estimation to support trustworthy and effective human–machine interaction. However, achieving such consistency remains challenging due to the subjective and context-dependent nature of affective data. In particular, individuals differ in expressivity, behaviour, and annotation patterns, while situational factors introduce additional variability in the observed signals~\cite{makantasis2022invariant,barthet2023knowing,hayat2022modeling}. As a result, affective models trained on pooled datasets often achieve satisfactory average performance but exhibit high variability in prediction error across participants, leading to unstable behaviour in real-world deployment~\cite{shoer2025learning}. From a modelling perspective, this instability arises from distributional shifts across participants, including differences in feature distributions, input–output relationships, and annotation reliability~\cite{krueger2021out,tamang2025handling}.

At the same time, deployment in pervasive environments necessitates compact and efficient models. Model pruning techniques address this requirement by reducing computational and energy costs~\cite{han2015learning}. However, most existing pruning approaches optimise primarily for sparsity or mean accuracy, without considering how parameter removal affects cross-user stability~\cite{vadera2022methods,cheng2024survey}. To address this gap, we propose Variance-Regularised Pruning (VR), a post-training pruning framework that explicitly couples model efficiency with robustness across participants. VR evaluates each connection based on its contribution to both the mean prediction error and its variance across participant-defined environments, removing parameters with minimal joint influence. The approach is inspired by Variance Risk Extrapolation (V-REx)~\cite{krueger2021out}, extending invariance principles to model sparsification without requiring architectural changes or retraining.

We evaluate the proposed variance-regularised pruning framework on the AGAIN dataset, which consists of affect-eliciting interactive scenarios annotated for continuous arousal~\cite{melhart2022arousal}. Experimental results show that the method achieves up to 80\% sparsity at the connection level while maintaining near-baseline Concordance Correlation Coefficient (CCC) performance. These findings indicate that incorporating variance-aware criteria into pruning enables the construction of compact yet robust affective models, supporting reliable operation under real-world deployment constraints. This paper makes three main contributions. First, it introduces a variance-regularised pruning framework tailored to affective computing. Second, it demonstrates that connection-level pruning guided by cross-participant stability can preserve predictive performance under high sparsity. Third, it shows that the proposed approach supports efficient and dependable affect estimation in pervasive, user-centred interactive environments.

\section{Related Work}

\textbf{}{Model Compression and Pruning:}
Model compression techniques aim to reduce computational cost while preserving predictive performance, enabling deployment in resource-constrained environments. Early pruning methods such as Optimal Brain Damage and Optimal Brain Surgeon removed parameters with low saliency based on second-order loss approximations~\cite{lecun1989optimal,hassibi1993optimal}. Later approaches popularised magnitude-based pruning~\cite{han2015learning}, while structured pruning removed entire neurons or filters to achieve hardware-friendly sparsity patterns~\cite{li2017pruning,he2017channel}. More recent work has explored sensitivity-based criteria~\cite{molchanov2019importance}, single-shot pruning~\cite{lee2018snip}, and sparse subnetwork discovery~\cite{frankle2018lottery,tanaka2020pruning}. While effective for improving efficiency, these methods primarily optimise for sparsity or aggregate accuracy and do not explicitly consider how pruning affects performance stability across users or environments. In contrast, our work introduces a variance-regularised pruning criterion that explicitly accounts for cross-participant variability, aligning model compression with robustness requirements in real-world interactive systems.

\textbf{Robust Learning under Distribution Shift:}
Robust learning under distributional shift has become increasingly important as machine learning systems move beyond controlled settings. Invariant Risk Minimisation (IRM)~\cite{arjovsky2019invariant} and Variance Risk Extrapolation (V-REx)~\cite{krueger2021out} promote predictors that perform consistently across environments by penalising environment-specific variability. Related approaches such as Group Distributionally Robust Optimisation~\cite{sagawa2019distributionally} similarly emphasise worst-case or underperforming environments to improve fairness and stability. In affective computing, each participant can be viewed as a distinct environment due to differences in expressivity, perception, and annotation behaviour. Prior work has addressed this challenge through domain adaptation and meta-learning~\cite{gideon2019improving,pinitas2024across}, often requiring retraining or specialised optimisation procedures. The present work differs by incorporating invariance principles into the \emph{post-training pruning stage}, enabling robustness-aware model compression without modifying the training process.

\textbf{Efficient Affective Computing in Interactive Systems:}
Affective computing systems deployed in interactive environments must balance efficiency with reliable performance across users. Benchmarks such as RECOLA~\cite{ringeval2013introducing} and Gamevibe~\cite{barthet2023knowing,barthet2024gamevibe} highlight substantial inter-participant variability arising from expressive diversity and annotation bias. Consequently, prior research has focused on cross-subject generalisation, domain adaptation, and lightweight model design~\cite{bai2025lightweight,makantasis2023lab}. In the context of interactive media and games, player modelling provides a natural testbed for affect-aware systems, linking emotional response to interaction dynamics~\cite{melhart2022arousal}. However, player behaviour and emotional reactivity remain highly individualised, often leading to unstable predictions across participants. Our work addresses this limitation by demonstrating that variance-aware pruning can simultaneously improve computational efficiency and cross-participant reliability, supporting the deployment of compact and dependable affective models in real-world interactive environments.

\section{Use Cases and Data Preprocessing}
This section provides an overview of the dataset and the preprocessing steps applied before experiments are performed, including a description of the collected gameplay features, the annotation protocol, and the data cleaning and normalisation procedures that prepare the signals for subsequent modelling.

\subsection{The AGAIN DATASET}

\begin{figure}[t]
  \centering
  \includegraphics[width=0.65\columnwidth]{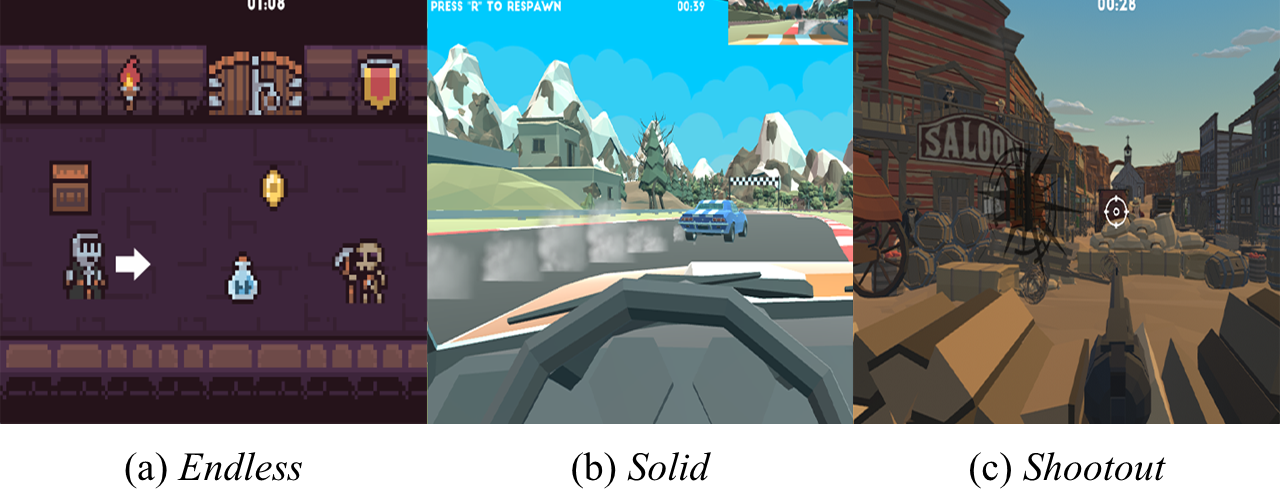}
  \caption{Example gameplay scenes from the three affect-eliciting games used in this study: (a) \textit{Endless} (platformer), (b) \textit{Solid} (racing), and (c) \textit{Shootout} (shooter), illustrating distinct interaction dynamics and sources of arousal.}
  \label{fig:all_games_again}
\end{figure}

The proposed methodology is evaluated using data from the AGAIN dataset~\cite{melhart2022arousal}, a corpus designed to study affect elicitation in fast-paced interactive environments. The dataset comprises short gameplay scenarios that emphasise immediate feedback, accessibility, and controlled interaction mechanics, making it suitable for investigating affect-aware systems operating under real-time and resource-constrained conditions. In this work, we focus on three representative games from AGAIN—\textit{Solid}, \textit{Shootout}, and \textit{Endless}—which span distinct interaction paradigms and sources of arousal commonly encountered in interactive systems. \textit{Solid} represents a racing scenario in which arousal arises from speed, control precision, and continuous visuomotor demands. \textit{Shootout} captures a shooter-style interaction where engagement is driven by aiming and shooting under increasing enemy density, inducing high attentional and motor load. \textit{Endless} is an infinite runner with automatic forward motion and increasing tempo, requiring rapid reactions to incoming obstacles. Together, these games provide complementary affective stimuli driven by speed, control, and perceptual load.

Data collection was conducted via online crowdsourcing. Following each gameplay session, participants annotated their perceived arousal using the PAGAN platform~\cite{melhart2019pagan}, adopting a stimulated recall protocol. This approach enables temporally aligned affect annotation while preserving ecological validity. Each gameplay session includes time-synchronised telemetry describing player behaviour and environmental context, such as position, velocity, health state, collisions, object counts, and interaction events. In addition to raw logs, the dataset provides structured telemetry features that capture both global gameplay dynamics and game-specific mechanics. These representations support fine-grained modelling of affective responses and are well-suited for evaluating efficient and robust affective models intended for deployment in real-world interactive environments.

\begin{figure*}[t!]
  \centering
  \includegraphics[width=1\textwidth]{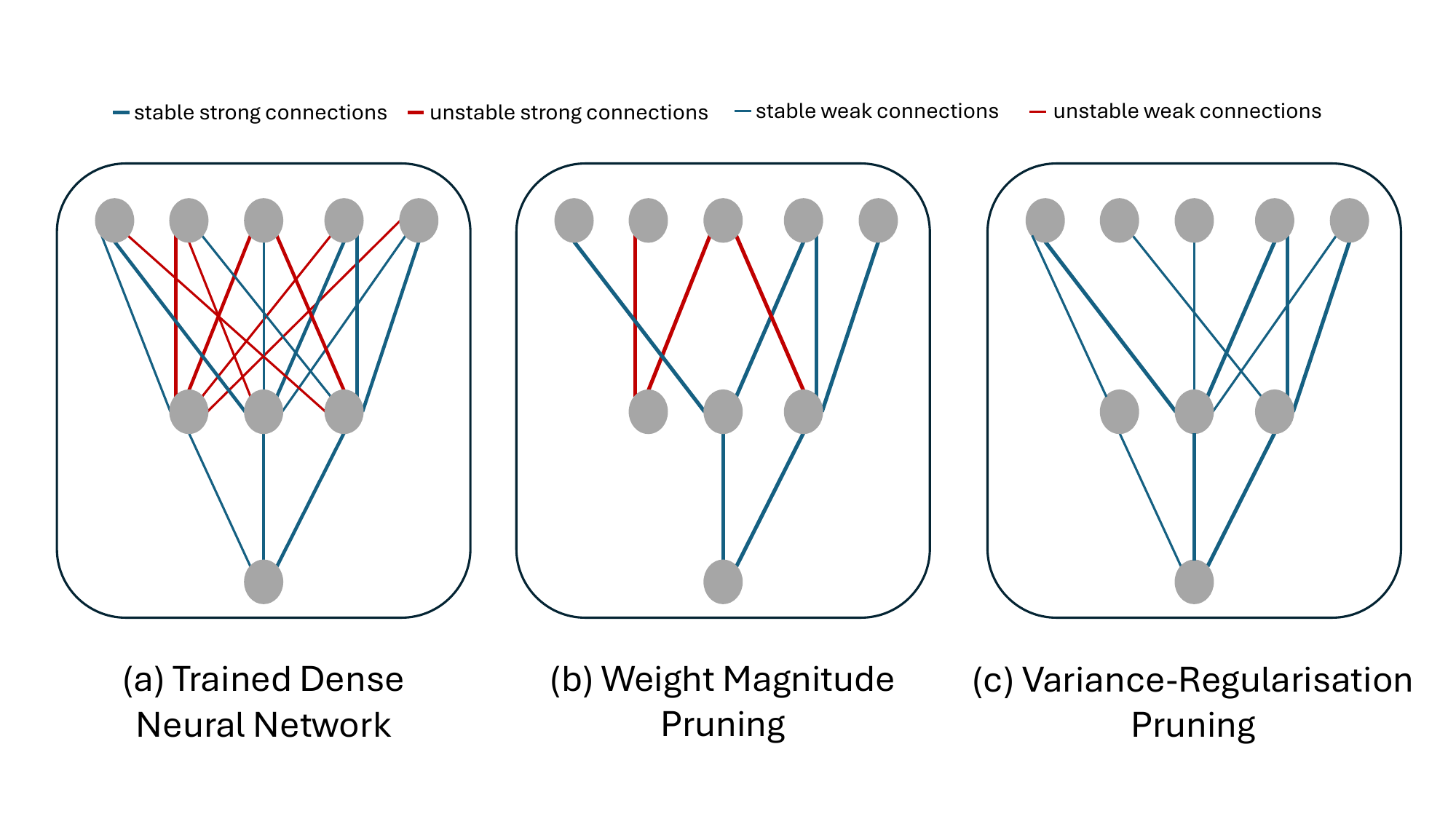}

  \caption{Illustration of the proposed variance-regularised pruning strategy compared with standard magnitude pruning.
(a) A trained dense neural network containing connections of varying strength and stability.
(b) Weight-magnitude pruning removes all weak connections, retaining only the strongest weights.
(c) Variance-regularised pruning preserves connections that are both important and contribute consistently across environments, leading to models that remain compact yet stable under distributional shifts.}
  \label{fig:method}
\end{figure*}

\subsection{Data Preprocessing}

The same preprocessing pipeline was applied across all games of the AGAIN dataset, following best practices in player modelling \cite{pinitas2025privileged,pinitas2024varying}, to prepare gameplay features and arousal annotations while preserving temporal dynamics. All features were segmented using a sliding window of three seconds with a one-second overlap between consecutive windows. Within each window, the feature vectors were concatenated to retain short-term temporal dependencies and contextual information across time. To ensure numerical stability and comparability across participants, all feature values were min–max normalised to the range $[0,1]$. Features with a variance lower than $0.01$ across the dataset were removed, as they provided little discriminative information.

For the arousal annotations, each trace was shifted backwards by one second to compensate for the average reaction delay of participants during the annotation process. Since the annotations are unbounded, all arousal traces were independently normalised to the range $[0,1]$. Within each temporal window, the arousal values were averaged to obtain a single representative value aligned with the corresponding feature segment. This preprocessing produced synchronised feature–target pairs suitable for model training and evaluation. It should be noted that the dimensionality of the input after preprocessing is 504 for \emph{Solid}, 346 for \emph{Shootout} and 465 for \emph{Endless}.

\section{Player Modelling Methodology}

In this study, the player modelling task is formulated as a supervised regression problem, where the objective is to predict continuous arousal levels from gameplay-derived features. Arousal represents the intensity of the player’s emotional activation and varies continuously over time, making regression a natural choice over discrete classification. Each training instance corresponds to a three-second gameplay segment, represented by the concatenated telemetry features extracted from the AGAIN dataset and paired with the average arousal value within the same time window.

\subsection{Baselines}

\subsubsection{Baseline Models}
Two fully connected artificial neural network (ANN) architectures were implemented as baseline models for arousal regression. Both models operate on the preprocessed feature representations extracted from the AGAIN dataset, where each input corresponds to the concatenated telemetry features within a three-second gameplay window. The networks map these fixed-length feature vectors to a single scalar output representing the predicted arousal value.

The first model, denoted as \emph{2-Layer}, consists of 1 hidden layer containing 256 neurons, followed by a linear output layer. This lightweight architecture serves as a compact baseline designed to evaluate the effectiveness of pruning strategies under limited capacity. The second model, referred to as \emph{5-Layer}, features a deeper hierarchy with hidden layers of sizes 768, 512, 384, and 192 neurons, respectively. This configuration allows for higher representational power and is intended to capture more complex, nonlinear relationships between gameplay features and arousal levels. Both architectures employ ReLU activations. All baseline models were trained using the same optimisation settings and loss objective (MSE), ensuring comparability between network sizes and subsequent pruning strategies.

\subsubsection{Baseline Pruning Methods}

To examine how model compression affects robustness in affective prediction, we consider baseline pruning methods applied after training. Pruning reduces computational and memory demands by removing parameters with limited influence on model output. Magnitude-based pruning serves as the foundation for most modern compression pipelines due to its simplicity, reproducibility, and interpretability~\cite{han2015learning,li2017pruning,he2017channel}.

Connection pruning (CP) eliminates individual weights while maintaining the original layer configuration. All methods operate post-training by applying magnitude thresholds to each weight matrix $\{\mathbf{W}^{(1)}, \ldots, \mathbf{W}^{(L)}\}$, producing unstructured sparsity patterns. This family of techniques isolates the effect of removing weak connections without altering architectural capacity.

\underline{CP-G (Global Magnitude Pruning).}
This method applies a single global threshold across all layers. All weights are pooled, and a quantile threshold corresponding to a target sparsity $s$ is computed. Connections with absolute values below this threshold are set to zero. Global pruning allows sparsity to emerge naturally according to each layer’s weight distribution, offering a consistent reference for compression performance. \underline{CP-L (Layer-wise Magnitude Pruning).}
In contrast, layer-wise pruning applies independent thresholds within each layer using predefined sparsity targets $s_l$. Each $\mathbf{W}^{(l)}$ is pruned according to its own distribution, preserving more parameters in sensitive layers and enforcing stronger sparsity in redundant ones. This approach enables finer control over sparsity allocation, avoiding layer imbalance often observed with global thresholds.

Neuron pruning introduces structured sparsity by removing entire hidden units, thereby reducing both the model’s parameter count and its representational capacity. \underline{NP-IN (Incoming-Norm Pruning).}
Each neuron’s importance is estimated via the L2 norm of its incoming weights. Only the top-$K$ neurons per layer with the largest norms are retained, removing weakly connected units. This criterion highlights neurons that contribute most to downstream activations from the input perspective. All neuron-level baselines are applied post hoc and reconstruct smaller multilayer perceptrons consistent with the selected neurons. Together, these baselines establish a comprehensive reference for comparing magnitude-based compression with the proposed variance-regularised pruning framework introduced in the next subsection.

\subsection{Variance-Regularised Connection Pruning }

We propose CP-VR a post-training pruning framework that produces \emph{compact yet reliable} affective models for deployment in interactive, user-centred environments. Unlike conventional magnitude- or sensitivity-based pruning, which optimises sparsity and average accuracy while ignoring user variability, our method explicitly couples compression with \emph{cross-participant stability}. Inspired by variance-based risk minimisation (e.g., V-REx~\cite{krueger2021out}), CP-VR favours connections whose influence is consistent across participant-defined environments.

To formalise the trade-off between predictive accuracy and cross-participant stability, we introduce a conceptual risk formulation that serves as a guiding principle for the pruning criterion. Let $\mathrm{MSE}_{\text{val}}$ denote the validation loss over all samples and $\mathrm{MSE}^{(g)}_{\text{val}}$ the loss computed within participant group $g$. We define
\begin{equation}
J \;=\; \mathrm{MSE}_{\text{val}} \;+\; \lambda_{\text{var}}\,\mathrm{Var}_{g}\!\big(\mathrm{MSE}^{(g)}_{\text{val}}\big),
\end{equation}
where $\lambda_{\text{var}}\!\ge\!0$ controls the balance between overall performance and stability across participants. Rather than minimising $J$ explicitly, we use it as a conceptual anchor to derive a variance-aware connection scoring rule that favours parameters contributing consistently across participant-defined environments.

Pruning is applied after training using a short calibration pass on held-out data. A single forward/backward pass estimates activation and gradient statistics $\mathbb{E}[a_i^2]$, $\mathbb{E}[g_j^2]$ and their group-specific counterparts $\mathbb{E}_g[a_i^2]$, $\mathbb{E}_g[g_j^2]$. For each weight $W_{ij}$, the average contribution is approximated by
\begin{equation}
\phi_{ij} \;=\; \tfrac{1}{2}\,\mathbb{E}[a_i^2]\,\mathbb{E}[g_j^2]\,W_{ij}^2,
\end{equation}
and the group-wise counterpart by
\begin{equation}
\phi^{(g)}_{ij} \;=\; \tfrac{1}{2}\,\mathbb{E}_g[a_i^2]\,\mathbb{E}_g[g_j^2]\,W_{ij}^2.
\end{equation}
We score each connection as
\begin{equation}
S_{ij} =\mu(\phi^{(g)}_{ij})+\lambda_{var}Var(\phi^{(g)}_{ij})
\end{equation}
Connections with the smallest $S_{ij}$ are pruned globally to reach the target sparsity, yielding unstructured sparsity without architectural changes. CP-VR thus removes weights that are both weak and \emph{inconsistent across participants}, retaining connections that support stable behaviour under user variability with minimal overhead (one calibration pass).

\section{Experiments}

\begin{figure}[tbh!]
    \centering
    \begin{subfigure}{0.45\linewidth}
        \centering
        \includegraphics[width=\linewidth]{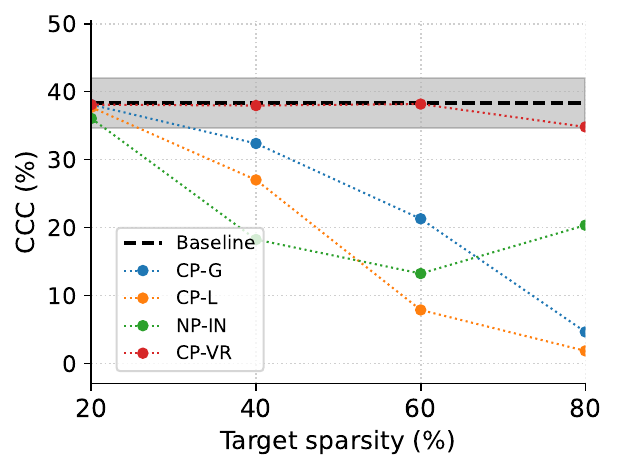}
        \caption{\textit{Endless, 2-layer}}
    \end{subfigure}\hfill
    \begin{subfigure}{0.45\linewidth}
        \centering
        \includegraphics[width=\linewidth]{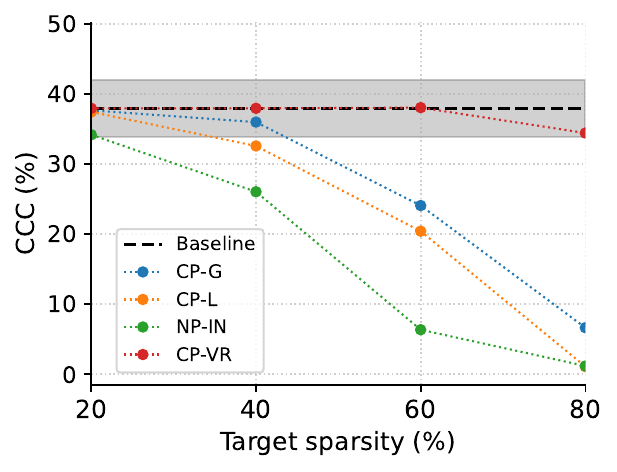}
        \caption{\textit{Endless, 5-layer}}
    \end{subfigure}\hfill
    \begin{subfigure}{0.45\linewidth}
        \centering
        \includegraphics[width=\linewidth]{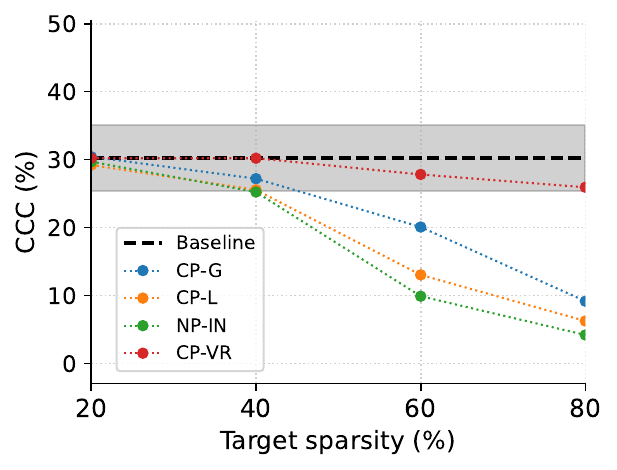}
        \caption{\textit{Solid, 2-layer}}
    \end{subfigure}\hfill
    \begin{subfigure}{0.45\linewidth}
        \centering
        \includegraphics[width=\linewidth]{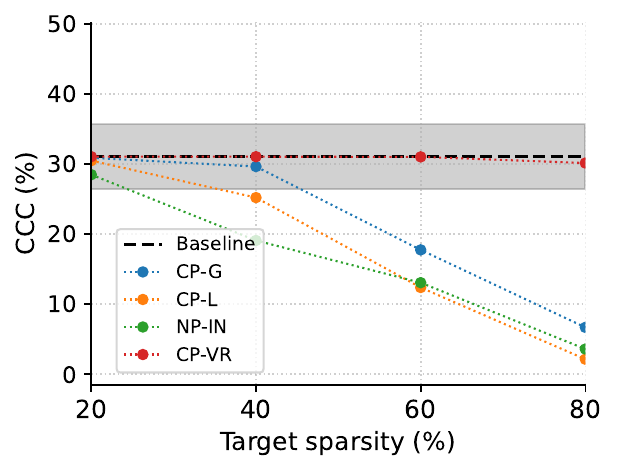}
        \caption{\textit{Solid,5-layer}}
    \end{subfigure}\hfill
    \begin{subfigure}{0.45\linewidth}
        \centering
        \includegraphics[width=\linewidth]{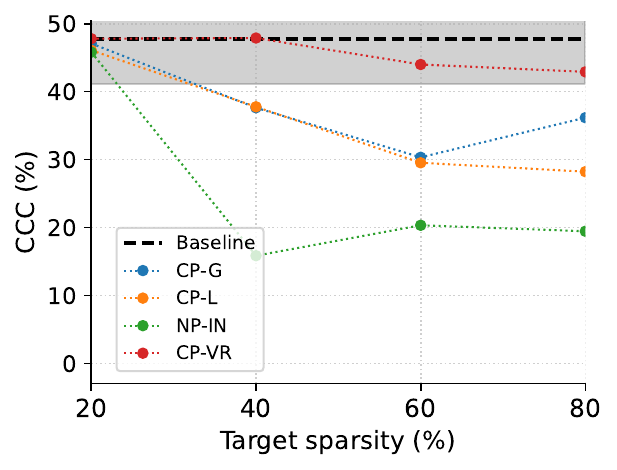}
        \caption{\textit{Shootout,2-layer}}
    \end{subfigure}\hfill
     \begin{subfigure}{0.45\linewidth}
        \centering
        \includegraphics[width=\linewidth]{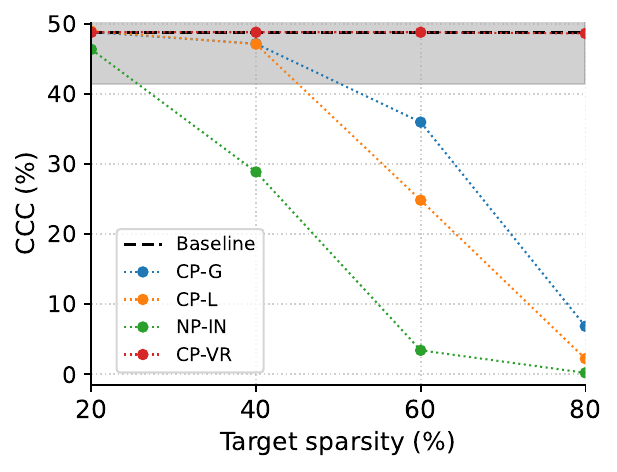}
        \caption{\textit{Shootout,5-layer}}
    \end{subfigure}
    \caption{Arousal prediction performance (CCC) under increasing sparsity, comparing the proposed variance-regularised pruning method against magnitude-based pruning baselines across the three games. A small 2-layer (left) and a larger 5-layer (right) network has been employed}
    \label{fig:res}
\end{figure}

This section evaluates the proposed pruning methods on the task of arousal regression using the AGAIN dataset. All experiments were designed to assess the relationship between model sparsity, predictive accuracy, and generalisation stability across players.

\subsection{Experimental Protocol}

All models were trained to predict continuous arousal using a regression formulation. To assess generalisation across users, the dataset was split by participant into non-overlapping training, validation, and test sets. Baseline networks were trained for 100 epochs using the Adam optimiser ($1\times10^{-3}$, $\beta_1=0.9$, $\beta_2=0.999$) with a batch size of 200, ReLU activations, and He initialisation. Mean squared error (MSE) was used for optimisation, with no additional regularisation. Models were subsequently pruned at the connection level with sparsity ranging from 0 to 80\%. Performance was evaluated primarily using the Concordance Correlation Coefficient (CCC), which captures agreement in continuous affective traces more effectively than MSE. Each experiment was repeated 15 times using different fixed random seeds. For variance-regularised pruning, the regularisation parameter was set to $\lambda_{\text{var}}=1$, with similar behaviour observed for nearby values.

\subsection{The Effects of Model Pruning}

Figure~\ref{fig:res} reports arousal prediction performance (CCC) under increasing sparsity for both a compact 2-layer network and a deeper 5-layer network, comparing the proposed variance-regularised pruning method (CP-VR) against magnitude-based pruning baselines across the three games. Overall, the results show that CP-VR consistently preserves predictive performance more effectively as sparsity increases, particularly at moderate to high pruning levels. For the 2-layer networks, magnitude-based pruning methods exhibit a gradual but steady degradation in performance as sparsity increases, with pronounced drops beyond 60\% sparsity. In contrast, CP-VR maintains near-baseline CCC values even at high sparsity levels, indicating that selectively retaining connections that are stable across participants leads to more reliable behaviour under aggressive compression. This trend is consistent across all three games, despite their differing interaction dynamics and arousal profiles.

The effect becomes more pronounced for the 5-layer networks, where increased model capacity amplifies sensitivity to pruning. While magnitude-based methods degrade rapidly—often exhibiting sharp performance drops at higher sparsity—the proposed method demonstrates substantially improved robustness, maintaining stable CCC values across a wide sparsity range. This suggests that variance-aware pruning is particularly beneficial for deeper architectures, where redundant yet unstable connections are more prevalent. Across games, the benefits of CP-VR are most evident in \textit{Shootout}, which induces high arousal variability and strong participant-specific effects. In this setting, CP-VR retains stable performance even when baseline methods fail, highlighting the importance of accounting for cross-participant variability during compression. Results for \textit{Endless} and \textit{Solid} show similar, though slightly less pronounced, trends, reflecting differences in interaction tempo and control demands. Taken together, these results demonstrate that variance-regularised connection pruning enables substantial model compression while preserving predictive accuracy and stability across users. This property is essential for affect-aware systems intended for deployment in real-world interactive environments, where computational constraints and user variability must be addressed simultaneously.

\section{Discussion and Conclusions}

This work introduced a variance-regularised pruning framework for compressing affective models while preserving reliable behaviour across users. Motivated by the high inter-participant variability inherent in affective data, the proposed approach extends conventional magnitude-based pruning by explicitly accounting for how parameter removal affects prediction stability across participant-defined environments. Rather than optimising sparsity alone, the framework favours retaining connections that contribute consistently across users, enabling compression aligned with robustness. Experiments on the AGAIN dataset demonstrated that variance-regularised \emph{connection-level} pruning achieves substantial compression—up to 80\% sparsity—while maintaining near-baseline Concordance Correlation Coefficient (CCC) performance across different games and network sizes. Compared to magnitude-based baselines, the proposed method degrades more gracefully under increasing sparsity and remains effective without retraining, highlighting its suitability for deployment in resource-constrained interactive systems.

Beyond performance gains, the proposed framework offers practical advantages. Pruning is applied post-training and requires only a short calibration pass, making it lightweight and architecture-agnostic. While this study focused on fully connected networks for arousal regression, the formulation is general and can be extended to other architectures and affective tasks. The reliance on participant identifiers to define environments represents a limitation, and future work may explore alternative grouping strategies or integrate variance-aware objectives directly into training. Overall, this work positions pruning not only as a tool for efficiency but also as a mechanism for improving robustness in affective computing. By coupling compression with variance-aware evaluation, the proposed approach provides a practical step toward efficient, reliable, and deployable affect-aware systems for real-world interactive environments.

\subsubsection{Ethical Impact Statement}
This work investigates affective modelling using anonymised interaction data collected with informed consent. All data used in the study were handled in accordance with applicable ethical guidelines, and no personally identifiable information is included. The proposed methods focus on improving the efficiency and reliability of affect-aware systems and do not introduce new forms of user profiling or surveillance. We do not anticipate significant risks of misuse, privacy violation, or discriminatory impact arising from this work.

%
%
%
 \bibliographystyle{splncs04}
 \bibliography{refs_new}

@article{lecun1989optimal,
  title={Optimal brain damage},
  author={LeCun, Yann and Denker, John and Solla, Sara},
  journal={Advances in neural information processing systems},
  volume={2},
  year={1989}
}

@inproceedings{hassibi1993optimal,
  title={Optimal brain surgeon and general network pruning},
  author={Hassibi, Babak and Stork, David G and Wolff, Gregory J},
  booktitle={IEEE international conference on neural networks},
  pages={293--299},
  year={1993},
  organization={IEEE}
}

@article{han2015learning,
  title={Learning both weights and connections for efficient neural network},
  author={Han, Song and Pool, Jeff and Tran, John and Dally, William},
  journal={Advances in neural information processing systems},
  volume={28},
  year={2015}
}

@inproceedings{he2017channel,
  title={Channel pruning for accelerating very deep neural networks},
  author={He, Yihui and Zhang, Xiangyu and Sun, Jian},
  booktitle={Proceedings of the IEEE international conference on computer vision},
  pages={1389--1397},
  year={2017}
}

@inproceedings{molchanov2019importance,
  title={Importance estimation for neural network pruning},
  author={Molchanov, Pavlo and Mallya, Arun and Tyree, Stephen and Frosio, Iuri and Kautz, Jan},
  booktitle={Proceedings of the IEEE/CVF conference on computer vision and pattern recognition},
  pages={11264--11272},
  year={2019}
}

@article{lee2018snip,
  title={Snip: Single-shot network pruning based on connection sensitivity},
  author={Lee, Namhoon and Ajanthan, Thalaiyasingam and Torr, Philip HS},
  journal={arXiv preprint arXiv:1810.02340},
  year={2018}
}

@article{tanaka2020pruning,
  title={Pruning neural networks without any data by iteratively conserving synaptic flow},
  author={Tanaka, Hidenori and Kunin, Daniel and Yamins, Daniel L and Ganguli, Surya},
  journal={Advances in neural information processing systems},
  volume={33},
  pages={6377--6389},
  year={2020}
}

@article{frankle2018lottery,
  title={The lottery ticket hypothesis: Finding sparse, trainable neural networks},
  author={Frankle, Jonathan and Carbin, Michael},
  journal={arXiv preprint arXiv:1803.03635},
  year={2018}
}

@article{arjovsky2019invariant,
  title={Invariant risk minimization},
  author={Arjovsky, Martin and Bottou, L{\'e}on and Gulrajani, Ishaan and Lopez-Paz, David},
  journal={arXiv preprint arXiv:1907.02893},
  year={2019}
}

@inproceedings{krueger2021out,
  title={Out-of-distribution generalization via risk extrapolation (rex)},
  author={Krueger, David and Caballero, Ethan and Jacobsen, Joern-Henrik and Zhang, Amy and Binas, Jonathan and Zhang, Dinghuai and Le Priol, Remi and Courville, Aaron},
  booktitle={International conference on machine learning},
  pages={5815--5826},
  year={2021},
  organization={PMLR}
}

@article{sagawa2019distributionally,
  title={Distributionally robust neural networks for group shifts: On the importance of regularization for worst-case generalization},
  author={Sagawa, Shiori and Koh, Pang Wei and Hashimoto, Tatsunori B and Liang, Percy},
  journal={arXiv preprint arXiv:1911.08731},
  year={2019}
}

@article{gideon2019improving,
  title={Improving cross-corpus speech emotion recognition with adversarial discriminative domain generalization (ADDoG)},
  author={Gideon, John and McInnis, Melvin G and Provost, Emily Mower},
  journal={IEEE Transactions on Affective Computing},
  volume={12},
  number={4},
  pages={1055--1068},
  year={2019},
  publisher={IEEE}
}

@inproceedings{pinitas2024across,
  title={Across-game engagement modelling via few-shot learning},
  author={Pinitas, Kosmas and Makantasis, Konstantinos and Yannakakis, Georgios N},
  booktitle={European Conference on Computer Vision},
  pages={390--406},
  year={2024},
  organization={Springer}
}

@inproceedings{makantasis2022invariant,
  title={The invariant ground truth of affect},
  author={Makantasis, Konstantinos and Pinitas, Kosmas and Liapis, Antonios and Yannakakis, Georgios N},
  booktitle={2022 10th International Conference on Affective Computing and Intelligent Interaction Workshops and Demos (ACIIW)},
  pages={1--8},
  year={2022},
  organization={IEEE}
}

@inproceedings{ringeval2013introducing,
  title={Introducing the RECOLA multimodal corpus of remote collaborative and affective interactions},
  author={Ringeval, Fabien and Sonderegger, Andreas and Sauer, Juergen and Lalanne, Denis},
  booktitle={2013 10th IEEE international conference and workshops on automatic face and gesture recognition (FG)},
  pages={1--8},
  year={2013},
  organization={IEEE}
}

@inproceedings{li2017pruning,
  title     = {Pruning Filters for Efficient ConvNets},
  author    = {Li, Hao and Kadav, Asim and Durdanovic, Igor and Samet, Hanan and Graf, Hans Peter},
  booktitle = {International Conference on Learning Representations (ICLR)},
  year      = {2017},
  url       = {https://arxiv.org/abs/1608.08710}
}

@article{melhart2022arousal,
  title={The arousal video game annotation (AGAIN) dataset},
  author={Melhart, David and Liapis, Antonios and Yannakakis, Georgios N},
  journal={IEEE Transactions on Affective Computing},
  volume={13},
  number={4},
  pages={2171--2184},
  year={2022},
  publisher={IEEE}
}

@inproceedings{barthet2023knowing,
  title={Knowing your annotator: Rapidly testing the reliability of affect annotation},
  author={Barthet, Matthew and Trivedi, Chintan and Pinitas, Kosmas and Xylakis, Emmanouil and Makantasis, Konstantinos and Liapis, Antonios and Yannakakis, Georgios N},
  booktitle={2023 11th International Conference on Affective Computing and Intelligent Interaction Workshops and Demos (ACIIW)},
  pages={1--8},
  year={2023},
  organization={IEEE}
}

@article{hayat2022modeling,
  title={Modeling subjective affect annotations with multi-task learning},
  author={Hayat, Hassan and Ventura, Carles and Lapedriza, Agata},
  journal={Sensors},
  volume={22},
  number={14},
  pages={5245},
  year={2022},
  publisher={MDPI}
}

@article{shoer2025learning,
  title={Learning Annotation Consensus for Continuous Emotion Recognition},
  author={Shoer, Ibrahim and Erzin, Engin},
  journal={arXiv preprint arXiv:2505.21196},
  year={2025}
}

@article{bai2025lightweight,
  title={Lightweight emotion analysis solution using tiny machine learning for portable devices},
  author={Bai, Maocheng and Yu, Xiaosheng},
  journal={Computers and Electrical Engineering},
  volume={123},
  pages={110038},
  year={2025},
  publisher={Elsevier}
}

@inproceedings{melhart2019pagan,
  title={PAGAN: Video affect annotation made easy},
  author={Melhart, David and Liapis, Antonios and Yannakakis, Georgios N},
  booktitle={2019 8th international conference on affective computing and intelligent interaction (acii)},
  pages={130--136},
  year={2019},
  organization={IEEE}
}

@inproceedings{picard1999affective,
  title={Affective computing for hci.},
  author={Picard, Rosalind W},
  booktitle={HCI (1)},
  pages={829--833},
  year={1999}
}

@article{yannakakis2023affective,
  title={Affective game computing: A survey},
  author={Yannakakis, Georgios N and Melhart, David},
  journal={Proceedings of the IEEE},
  volume={111},
  number={10},
  pages={1423--1444},
  year={2023},
  publisher={IEEE}
}

@article{tamang2025handling,
  title={Handling Out-of-Distribution Data: A Survey},
  author={Tamang, Lakpa and Bouadjenek, Mohamed Reda and Dazeley, Richard and Aryal, Sunil},
  journal={IEEE Transactions on Knowledge and Data Engineering},
  year={2025},
  publisher={IEEE}
}

@article{cheng2024survey,
  title={A survey on deep neural network pruning: Taxonomy, comparison, analysis, and recommendations},
  author={Cheng, Hongrong and Zhang, Miao and Shi, Javen Qinfeng},
  journal={IEEE Transactions on Pattern Analysis and Machine Intelligence},
  year={2024},
  publisher={IEEE}
}

@article{vadera2022methods,
  title={Methods for pruning deep neural networks},
  author={Vadera, Sunil and Ameen, Salem},
  journal={Ieee Access},
  volume={10},
  pages={63280--63300},
  year={2022},
  publisher={IEEE}
}

@article{barthet2024gamevibe,
  title={GameVibe: a multimodal affective game corpus},
  author={Barthet, Matthew and Kaselimi, Maria and Pinitas, Kosmas and Makantasis, Konstantinos and Liapis, Antonios and Yannakakis, Georgios N},
  journal={Scientific data},
  volume={11},
  number={1},
  pages={1306},
  year={2024},
  publisher={Nature Publishing Group UK London}
}

@article{makantasis2023lab,
  title={From the lab to the wild: Affect modeling via privileged information},
  author={Makantasis, Konstantinos and Pinitas, Kosmas and Liapis, Antonios and Yannakakis, Georgios N},
  journal={IEEE Transactions on Affective Computing},
  volume={15},
  number={2},
  pages={380--392},
  year={2023},
  publisher={IEEE}
}

@inproceedings{pinitas2025privileged,
  title={Privileged Contrastive Pretraining for Multimodal Affect Modelling},
  author={Pinitas, Kosmas and Makantasis, Konstantinos and Yannakakis, Georgios},
  booktitle={Proceedings of the 27th International Conference on Multimodal Interaction},
  pages={317--325},
  year={2025}
}

@inproceedings{pinitas2024varying,
  title={Varying the context to advance affect modelling: A study on game engagement prediction},
  author={Pinitas, Kosmas and Rasajski, Nemanja and Barthet, Matthew and Kaselimi, Maria and Makantasis, Konstantinos and Liapis, Antonios and Yannakakis, Georgios N},
  booktitle={2024 12th International Conference on Affective Computing and Intelligent Interaction (ACII)},
  pages={194--202},
  year={2024},
  organization={IEEE}
}

\end{document}